\documentclass{llncs}

\usepackage{amsmath}
\usepackage{amssymb}
\usepackage{graphicx}
\usepackage{hyperref}
\DeclareMathOperator*{\argmax}{arg\,max}
\begin{document}

\title{Prediction of Horizontal Data Partitioning Through Query Execution Cost Estimation}

\author{Nino Arsov \inst{1}, Goran Velinov \inst{1}, Aleksandar S. Dimovski \inst{2}, Bojana Koteska \inst{1}, Dragan Sahpaski \inst{1}, Margita Kon-Popovska \inst{1}}
\institute{{Faculty of Computer Science and Engineering, Ss. Cyril and Methodius Universiy\\Rugjer Boshkovikj 16\\Skopje,Macedonia},{IT University of Copenhagen\\Rued Langgaards Vej 7, 2300\\Copenhagen,Denmark}}

\maketitle
\begin{abstract}
The excessively increased volume of data in modern data management systems demands an improved system performance, frequently provided by data distribution, system scalability and performance optimization techniques. Optimized horizontal data partitioning has a significant influence of distributed data management systems.
An optimally partitioned schema found in the early phase of logical database design without loading of real data in the system and its adaptation to changes of business environment are very important for a successful implementation, system scalability and performance improvement.
\\	
In this paper we present a novel approach for finding an optimal horizontally partitioned schema that manifests a minimal total execution cost of a given database workload. Our approach is based on a formal model that enables abstraction of the predicates in the workload queries, and are subsequently used to define all relational fragments. This approach has predictive features acquired by simulation of horizontal partitioning, without loading any data into the partitions, but instead, altering the statistics in the database catalogs. We define an optimization problem and employ a genetic algorithm (GA) to find an approximately optimal horizontally partitioned schema. The solutions to the optimization problem are evaluated using PostgreSQL's query optimizer.
The initial experimental evaluation of our approach confirms its efficiency and correctness, and the numbers imply that the approach is effective in reducing the workload execution cost.
\end{abstract}


\section{Introduction}
\label{sec:introduction}

The focus of this paper is the prediction of an (approximately) optimal horizontally partitioned schema through simulation of horizontal data partitioning, which is performed by altering the database catalogs (statistics). We use the optimizer cost model to estimate the execution cost of a given workload. Our motivation is based on four key factors.

First, although the problem of (optimal) horizontal partitioning is well studied, to find a partitioned schema that minimizes the workload execution cost is still a challenging problem~\cite{DBLP:conf/birte/JindalD11,DBLP:dblpconf/dexa/LirozGistauAPPV12}.
The idea behind it is to distribute the rows of a relational table across the nodes of a cluster so that they can be processed in parallel~\cite{Stonebraker:2010:MPD:1629175.1629197}.
In this way, the system performance for a workload can be significantly improved, since parallel processing of data becomes possible by placing the tuples where they are most frequently accessed. 
In effect, the workload performance can scale to larger volumes of data. The work on this problem has been extensive~\cite{61401488,bellatrechereferential,DBLP:dblpconf/dexa/LirozGistauAPPV12,Ozsu2011principles}.
The ideas were then adapted to the setting of a data warehouse, where significant performance improvements are possible due to the size of the fact table in a star or a snowflake schema. However, the optimal partitioning problem is NP-hard and static (non-adaptive) solutions are not suitable for dynamically changing workloads~\cite{bellatreche2013horizontal}. 
Automated data partitioning methods in parallel database systems have also been proposed, since partitioning can have a positive impact on the system performance~\cite{Nehme:2011:APD:1989323.1989444,Pavlo:2012:SAD:2213836.2213844}.
Our motivation is to propose an efficient method that addresses the problem of optimal horizontal data warehouse partitioning.

Second, in the early phase of design of analytical systems it is very important to predict an optimal horizontally partitioned schema that minimizes the workload execution cost.
Also in the case of changes to some of the factors that affect the system's performance, it is
important to find a new optimal design (partitioned schema) with minimal reallocation cost.
This problem is partially addressed by on-line data re-balancing, which ensures storage balance at all times, even after insertion/deletion~\cite{Ganesan:2004:OBR:1316689.1316729}. The approach is adaptive, but it does not ensure optimal performance.

Third, the inevitable increase of the volume of data in analytical systems introduces the need of improved system scalability, elasticity and adaptivity in order to avoid additional costs related to new hardware configurations and costly reorganization or a complete logical redesign. Optimal on-line horizontal data (re)partitioning can help avoid these costs by significantly improving the scalability of existing hardware configurations of analytical systems. On-line analytical processing (OLAP) systems used in data-driven decision making have to be elastic, i.e. adaptive to changing workloads over time to meet the requirements of the business environment. These characteristics related to big data can be obtained by horizontal partitioning based on a given workload.

Finally, there is extensive research and good novel results in the field of query execution cost estimation
and database optimizer model improvements. The cost models used by query optimizers are challenged by statistical machine learning approaches, but if properly calibrated, they can be highly accurate and precisely reflect the real execution cost of queries~\cite{wu13prediction}. Therefore, the query optimizer models become even more promising
and our approach is more applicable on real systems~\cite{Herodotou:2010:XSQ:1920841.1920984,Akdere:2012:LQP:2310257.2310339}.

In an endeavor to accomplish these objectives, we develop a new approach for automatic generation of an optimal/good horizontally partitioned schema. The main characteristics of our approach 
for automatic generation of optimal/good horizontally partitioned schema are the following:
\begin{itemize}
	\item The approach predicts the execution cost of the workload without loading real data in the database,
	only by changing the statistics of the database system.
	\item The approach finds an approximately optimal/good partitioned schema of a data warehouse for a given representative workload. We use a GA to find an approximately optimal horizontally partitioned schema.
	%
	\item The approach is based on a formal model for horizontal partitioning by predicate abstraction and uses a real query optimizer to estimate the total execution cost of a given workload. It is applied to PostgreSQL's query optimizer.
\end{itemize}

Relational data warehouses often contain large relations (fact relations or fact tables) and require techniques both for managing (maintaining) these large relations and for providing good workload performance across these large relations. The space of possible physical partitioning schema alternatives that need to be considered is very large and grows exponentially with respect to the number of range predicates used for range partitioning.

To address the optimization problem, we first choose a set of predicates to horizontally partition
some (or all) of the dimension relations of a data warehouse with a star schema. The predicates over a relation can involve one or more attributes of that relation. 
Then we split the fact relation by using the predicates specified on dimension relations. This creates a number of sub-star fragments of the data
warehouse we consider, where each sub-star fragment consists of a partition of the fact table
and corresponding to it partitions of dimension relations.
Then we find a suitable solution which minimizes the query cost.
To validate the efficiency of our approach, the experiments are conducted using
the Star Schema Benchmark (SSB) dataset, an adapted  data warehouse variation of TPC-H dataset~\cite{o2007star},
and the JGAP genetic algorithm package is used to implement the GA~\cite{JGAP}. Our approach does not guarantee the best possible partitioning, but the experimental results suggest that it produces good solutions in practice. In this paper we present a proof of concept of our approach to optimal horizontal partitioning.

The paper is organized as follows. We discuss related work in Section~\ref{background}, where
we describe an existing formal model for horizontal partitioning of relations and data warehouses.
A procedure for simulation of horizontal partitioning of relations is presented in Section~\ref{simulation}.
In Section~\ref{optimal}, the optimization problem is defined and a genetic algorithm addressing it is described.
We present experimental results in Section~\ref{experiments}. Finally, in Section~\ref{conclusion}, we conclude and discuss future work. On a short notice, the terms \textit{fragmentation} and \textit{partitioning} are used interchangeably.

\section{Background Work}
\label{background}

Further, this section gives an overview of a formal model for horizontal partitioning of relations and data warehouses
based on predicate abstraction, as defined in~\cite{dimovski2010horizontal}. 

Let $R$ be a relation, and $A_1, ..., A_n$ be its attributes with the corresponding domains $Dom(A_1), ..., Dom(A_n)$.
The set of all predicates over a relation $R$ is defined by:
\[
\phi ::= p \, | \, \neg \phi \, | \, \phi_1 \land \phi_2 \, | \, \phi_1 \lor \phi_2
\]
where $p$ is an \emph{atomic predicate}, i.e.\ relationship among attributes and constants of a relation.

We define a two-phase \emph{horizontal partitioning} as a pair $(R,\phi)$, where $R$ is a relation and $\phi$ is a predicate.
It splits $R$ into at most 2 fragments of $R$ with the identical structure, one per each truth value of $\phi$,
i.e. we have:
\[ \begin{array}{l}
R_{(0)} = \{ t \in R \, | \, t \vDash \neg \phi \} \\
R_{(1)} = \{ t \in R \, | \, t \vDash \phi \} \\
\end{array} \]
where the first fragment $R_{(0)}$ includes all tuples $t$ of $R$ which do not satisfy $\phi$,
the second fragment $R_{(1)}$ includes all tuples $t$ of $R$ which  satisfy $\phi$.
It is possible one of the fragments to be empty if all tuples of $R$ either satisfy or do not satisfy $\phi$.
In the second phase, it is allowed to merge some of the fragments obtained previously.
In this case, that can be done by discarding the predicate $\phi$.
\[
R_{(2)} = \{ t \in R \, | \, true \} = R
\]
We can apply horizontal partitioning using a predicate $\phi_2$
to each of the fragments obtained by a partitioning $(R,\phi_1)$, denoted as $(R,\phi_1,\phi_2)$,
thus obtaining at most 4 fragments of $R$.
They are denoted as: $R_{(0,0)}$, $R_{(0,1)}$, $R_{(1,0)}$, and $R_{(1,1)}$.
In the second phase, we can also merge some of these fragments.
For example, $R_{(2,0)} = \{ t \in R \, | \, t \vDash \neg \phi_2 \}$
is obtained by merging $R_{(0,0)}$ and $R_{(1,0)}$, while
$R_{(1,2)} = \{ t \in R \, | \, t \vDash \phi_1 \}$ by merging $R_{(1,0)}$ and $R_{(1,1)}$.

This is called embedded horizontal partitioning, and it can be applied with an arbitrary number
of predicates $m$,
such that in each level a new predicate is applied to the obtained fragments.
Embedded horizontal partitioning of a relation $R$ with $m$ predicates is denoted as
$(R,\phi_1,\phi_2, ...,\phi_m)$, and it can split the initial relation $R$
into at most $2^m$ fragments, denoted as:
\[
R_{(v_1, \ldots, v_m)} = \{ t \, | \, t \vDash v_1 \cdot \phi_1 \, \land \ldots \land \, v_m \cdot \phi_m \}
\]
where $v_i \in \{0,1\}$, $1 \leq i \leq m$, and $0 \cdot \phi = \neg \phi$, $1 \cdot \phi = \phi$.
Again, in the second phase of partitioning we can decide to merge some of
the fragments by discarding some of the predicates.
In this case, we have that:
\[
R_{(v_1, \ldots, v_m)} = \{ t \, | \, t \vDash v_1 \cdot \phi_1 \, \land \ldots \land \, v_m \cdot \phi_m \}
\]
where $v_i \in \{0,1,2\}$, $1 \leq i \leq m$, and $0 \cdot \phi = \neg \phi$, $1 \cdot \phi = \phi$, $2 \cdot \phi = true$.
An index table with $2^m$ entries representing all possible bit-vectors of length $m$ can be formed:
\[
\{ (v_1, \ldots, v_m) \, | \, v_i \in \{0,1\}, \, i=1, \ldots, m \}
\]
An index entry $(v_1, \ldots, v_m)$ from the index table points to the fragment $R_{(v_1, \ldots, v_m)}$.
If some fragment is empty, then there will be no pointer to it.
If two fragments are merged, it is possible that two entries point to the same fragment.
Then local index tables can be created on each of the fragments.

Derived Horizontal Partitioning is defined on a relation which refers to
another relation by using its primary key as reference.
Let $R=(A_1, \ldots, A_n)$ and $S=(B_1, \ldots, B_m)$ be relations,
such that $S$ contains a foreign key referring to a primary key of $R$.
Given a horizontal partitioning of $R$
into $R_1, \ldots, R_k$, this induces the derived horizontal fragmentation of $S$
into $k$ fragments:
\[
S_l = S \ltimes R_l, \, l=1, \ldots, k
\]
where the semi-join operator $\ltimes$ is defined as\\
$S \ltimes R = \pi_{B_1, \ldots, B_m}(S \Join R)$,
i.e. the result is the set of all tuples in $S$ for which there is a tuple in $R$
that is equal on their common attributes.

Consider a relational data warehouse modeled by a star schema $(F,D_1, D_2, \ldots, D_k)$,
where $F$ is a fact relation and $D_1, \ldots, D_k$ are dimension relations.
Suppose that each dimension $D_i$
is horizontally partitioned by using a set of predicates $\{ \phi_{i,1}, \phi_{i,2}, \ldots \phi_{i,m_i} \}$
obtaining in such a way at most $2^{m_i}$ fragments $D_{i_{(v_{i,1}, \ldots, v_{i,m_i})}}$, where $1 \leq i \leq k$.
Then, the fact relation $F$ is partitioned using derived
horizontal partitioning in the following way:
\begin{flalign*}
&F_{(v_{1,1}, \ldots, v_{1,m_1}, \ldots, v_{k,m_{k}})} = \\
&= \big(F \ltimes D_{1_{(v_{1,1}, \ldots, v_{1,m_1})}} \ltimes \ldots
\ltimes D_{k_{(v_{k,1}, \ldots, v_{k,m_k})}} \big)
\end{flalign*}
where $v_{i,j} \in \{0,1\} $, for $1 \leq i \leq k$ and $1 \leq j \leq m_{i}$.
So we obtain at most $2^{\sum^{k}_{i=1}m_i}$ fragments of the fact relation.
Given a fact relation partition $F_{(v_{1,1}, \ldots, v_{1,m_1}, \ldots, v_{k,m_{k}})}$,
we can create the following sub-star schema fragment:
\[
(F_{(v_{1,1}, \ldots, v_{1,m_1}, \ldots, v_{k,m_{k}})},D_{1_{(v_{1,1}, \ldots, v_{1,m_1})}}, \ldots, D_{k_{(v_{k,1}, \ldots, v_{k,m_k})}})
\]
We can form a global index table with $2^{\sum^{k}_{i=1}m_i}$ entries representing all possible bit-vectors of length $\sum^{k}_{i=1}m_i$.
Each single entry $(v_{1,1}, \ldots, v_{1,m_1}, \ldots, v_{k,m_{k}})$ from the index points to the
sub-star schema created by dimension relations\\ $D_{i_{(v_{i,1}, \ldots, v_{i,m_i})}}$ for $1 \leq i \leq k$, and the corresponding fact sub-relation.
Then local index tables are created on each of the sub-star schemas.

\section{Prediction of Data Partitioning}
\label{prediction}

The extracted predicates from the workload are used for partitioning by predicate abstraction, which is optimized by a GA.
The quality of the solution is input-sensitive. There are two factors that cause this sensitivity. First, the execution cost is sensitive to the size of the table subregions (or partitions) being accessed. The size of each partition is determined by the selectivity of the predicates used for range partitioning over the non-partitioned data. Second, the breadth and depth of the search in the space of horizontally partitioned schemas are determined by the size of the population and the number of evolution generations, respectively.

We assume that a non-partitioned relational schema is given and that its statistical data exist in the system catalogs. The simulation of horizontal partitioning is performed by creating a new table for each partition, defined by a logical conjunction of one or more atomic predicates. Actual data are not loaded into these partitions, and obtaining highly accurate cost estimations from the query optimizer requires only the presence of statistical catalog data in a given DBMS. The execution cost estimation is inexpensive in terms of disk accesses, since it only requires a query execution plan (query tree), and, no queries need to be executed against actual data in the partitions.

As the section progresses further, three major statistical data structures are reviewed and the corresponding statistics generation methods are encompassed. For each column, we examine one-dimensional histograms, a list of most common values and their corresponding frequencies. Together with distinct values fractions, attribute widths and table metadata (cardinality in terms of both tuples and disk pages), they form an exhaustive basis for a functional query optimizer and query plan generation.


\subsection{Derived Partition Statistics}
\label{derived_stats}
 

 Let $R$ be a non-partitioned relation with attributes $A_1$, \dots, $A_n$. For each fragment $R_{(v_1, \dots, v_m)}$ obtained by partitioning the parent relation $R$ with a set of atomic predicates \{$\phi_1$, \dots, $\phi_m$\}.  The \textit{histogram} is a list of interval boundaries that split the values of the attribute into bins (buckets or groups) of approximately equal size; \textit{most common values} is a list of the most frequent non-null attribute values, and \textit{most common frequencies} array stores their frequencies. 


Three data structures are defined for an attribute $A$ of $R$.  Let $H=\{ b_1, \ldots, b_z \}$ be the one-dimensional \textit{histogram},
where $b_1,\dots,b_z$ are interval boundaries from $Dom(A)$ which define $z-1$ buckets,
$[b_1,b_2), [b_2,b_3), \dots, [b_{z-1},b_{z})$, such that the number of tuples from $R$
in each bucket is approximately the same.
Let $Val[\,]$ represent the array of $v$ \textit{most common values} from $Dom(A)$, and let $Fre[\,]$ store the \textit{most common frequencies},
such that $Fre[i]$ is the frequency of $Val[i]$, for $i = 1, \dots, v$. 

It is assumed that the subscript of each data structure represents its  relation $R$ (or fragment $R_{(v_1, \ldots, v_m)}$), and that they refer to a single attribute $A$ of $R$.
The values in $H_{R_{(v_1, \ldots, v_m)}}$ are chosen such that
\begin{flalign*}
& H_{R_{(v_1, \ldots, v_m)}} =\{ b_i \mid b_i \in H_R, i\in \{1,\ldots,z\}, \\
& \qquad \qquad \qquad b_i \vDash {\phi_1} \,\land\, \ldots \land\, b_i \vDash \phi_m \}
\end{flalign*}
The values in $Val_{R_{(v_1, \ldots, v_m)}}[\,]$ are chosen such that
\begin{flalign*}
& Val_{R_{(v_1, \ldots, v_m)}}[\,]=\{ Val[i] \mid Val[i] \in Val_R[\,], i \!\in\!\{1,\ldots,v\}, \\ 
& \qquad \qquad \qquad \quad Val[i]\vDash \phi_1 \land \ldots \land\, Val[i]\vDash \phi_m, \}
\end{flalign*}
The values for $Fre_{R_{(v_1, \ldots, v_m)}}[\,]$ are chosen such that\\ $Fre_{R_{(v_1, \ldots, v_m)}}[i]$ is the frequency of $Val_{R_{(v_1, \ldots, v_m)}}[i]$ in\\ $R_{(v_1, \ldots, v_m)}$, and $0<Fre_{R_{(v_1, \ldots, v_m)}}[i]\leq1$.

The width $w$ of an attribute is the average size in bytes of the data stored in each field of the corresponding column. If the column data is represented with a fixed-length type (such as 32-bit integer), its width in the fragment is equal the same value as in parent relation. If the column data is represented by a variable-length type (such as text), the width of the attribute has to be recalculated from the data. Copying the width value from the parent relation can lead to highly inaccurate estimates for a non-uniform, or skewed distribution of size of data in each field in the column.

The fraction of distinct values $\sigma$ is the ratio of the number of distinct values to the total number of records in an arbitrary column. Therefore, $\sigma=|Dom(A)|\,/\,|R|$.

Since all of these statistics can be easily estimated by querying the original data, which breaks down to loading data into every partition, time-efficient structures are needed to quickly derive all of the statistics for a partition. In the following part, we present two data structures that can be effectively employed for statistics derivation.

\subsection{Multidimensional Histograms of Finest Granularity}
\label{sec:multidim_hist}
A multidimensional histogram (MDH) provides a substantial part of statistical data necessary for estimating a partition's statistics. Histograms are a fundamental part of planning query execution, and are certainly ubiquitous across commercial systems; PostgreSQL supports only one dimensional histograms, while other systems, such as Oracle DB support multidimensional ones~\cite{OracleDB}. Generally, histogram granularity is referred to as the number of bins, but in the case of horizontal partitioning by predicate abstraction, it is referred to as the smallest range partition, i.e. the most constrained one, being defined by the largest number of workload predicates possible. This approach allows for flexibility, since the level of granularity is easy to adjust through aggregation of more granular records. Since a GA is the optimization method of choice (see Section~\ref{sec:introduction}) and any predicate can be used for partitioning, the histogram presented in this section is of finest granularity. It contains at most $2^{|\mathcal{P}|}$ records for the most detailed fragments of $R$, where $\mathcal{P}$ is the set of all extracted atomic predicates from the workload, for the relation $R$. These fragments are mutually exclusive, and each record in $\mathit{MDH}_R$ is a key-value pair, such that
\begin{flalign*}
\textit{MDH}_R=\{&\left((v_1,\ldots,v_{|\mathcal{P}|}), (T_1,\ldots,T_n)\right) \mid v_j\in\{0,1\},\\
& j=1,\ldots,|\mathcal{P}|\},
\end{flalign*}
where $T_k$ is a self-balancing AVL tree~\cite{sedgewick2002algorithms} that stores information for the $k$-th attribute $A_k$ of the fragment $R_{(v_1,\ldots,v_{|\mathcal{P}|})}$. Each node of the AVL tree $T_k$ is, also, a key-value pair $(d_k, c_{d_k}),\, d_k\in Dom(A_k)$, where $c_{d_{k}}$ denotes the number of occurrences of $d_{k}$ in the $k$-th column of $R_{(v_1,\ldots,v_{|\mathcal{P}|})}$. The AVL tree $T_k$ is ordered by the keys of its nodes. Balanced AVL trees guarantee $O(\log{|Dom(A_k)|})$ complexity for lookup. Hash algorithms are deliberately left out because of the size of the data. For brevity in the following part it is assumed that calculations refer to the statistics of a single attribute $A_k$ of $R$.

A record for any fragment $R_{(v_1,\ldots, v_m)}$, generated by partitioning $R$ using a subset of atomic predicates $\{\phi_1,\ldots, \phi_m\}\subseteq \mathcal{P}$, can be derived by merging the fragments in $\mathit{MDH}_R$ by keys that contain $(v_1,\ldots,v_m)$ as a subsequence, and aggregating their values (e.g summation). The multidimensional histograms for each relation $R$ from the non-partitioned schema are constructed prior to any statistics estimations.

First, the histogram $H_R$ and most common values array $Val_R[\,]$ are loaded into main memory, if they exist. The former is ordered, while the latter is sorted upon loading. Any value of $A_k$ can exist in either $H_R$ or $Val_R[\,]$, or both. To construct $H_{R_{(v_1, \ldots, v_m)}}$ or $Val_{R_{(v_1, \ldots, v_m)}}$, it is required to scan a list of all distinct values from $Dom(A_k)$ that satisfy the fragment's predicates \{$\phi_1,\ldots, \phi_m$\}. For each value in the list, binary search is used to check its existence in $H_R$ or $Val_R[\,]$ and it is added to  $H_{R_{(v_1, \ldots, v_m)}}$ or $Val_{R_{(v_1, \ldots, v_m)}}$, or both, respectively.

The most common frequencies array $Fre_{R_{(v_1,\ldots,v_m)}}[\,]$ is recomputed using $Val_{R_{(v_1,\ldots,v_m)}}[\,]$ and $\mathit{MDH}_R$. All keys of $\mathit{MDH}_R$ are scanned and their values are aggregated, such that for any key $(v_1,\ldots,v_{|\mathcal{P}|})$,
\[
Fre_{R_{(v_1,\ldots,v_m)}}[i]=\frac{\displaystyle\sum_{(v_1,\ldots,v_m)\subseteq (v_1,\ldots,v_{|\mathcal{P}|})}T_k(Val_{R_{(v_1,\ldots,v_m)}}[i])}{|R_{(v_1,\ldots,v_m)}|}.
\]

The average width $w_k$ of $A_k$ is copied from $R$'s statistics, if $A_k$ is represented with a fixed-length data type, or else, if a variable-length data type is used, it is recomputed by traversal of every AVL tree $T_k$ in records of $\mathit{MDH}_R$ where $(v_1,\ldots,v_m)\subseteq (v_1,\ldots,v_{|\mathcal{P}|})$.

\subsection{Roaring Bitmap Indexes for Fast Set Operations}
\label{sec:roaring_bitmap}
The estimation of statistics also requires set intersection operations, for which $\mathit{MDH}_R$ is inefficient. They can be optimized by bitmap indexes using bitwise operations implemented in the CPU hardware.
Since standard bitmap indexes can be inefficient in terms of memory and speed of bitmap operations, we use Roaring Bitmap, a two-level heterogeneous memory efficient compressed bitmap index with optimized bitmap operations~\cite{LemireRoaringBitmaps}.
Two flavors of Roaring Bitmap indexes are used to derive the fraction of distinct values $\sigma$ and the cardinality of a table, i.e. the number of tuples (rows).

To compute the fraction of distinct values $\sigma_{R_{(v_1,\ldots,v_m)}}$ of an attribute $A_k$ of the fragment $R_{(v_1,\ldots,v_m)}$, roaring bitmap indexes $B_{k\phi_j},\,\phi_j\in\{\phi_1,\ldots,\phi_m\}$  are used to avoid expensive set intersection operations of the AVL trees in $\mathit{MDH}_R$. First, the domain $Dom(A_k)$ of $A_k$ is used to encode each extracted atomic predicate $p\in\mathcal{P}$. Then, for any set of predicates \{$\phi_1,\ldots,\phi_m$\}, used for horizontal partitioning, the number of distinct values of $A_k$ in $R_{(v_1,\ldots,v_m)}$ can be quickly calculated as
\begin{equation}
|B_{k\{\phi_1,\ldots,\phi_m\}}^{(v_1,\ldots,v_m)}|=|B_{k\phi_1}^{v_1}\,\land\,\ldots\,\land\,B_{k\phi_m}^{v_m}|,
\label{eq:stadistinct_bitmap_index}
\end{equation}
where $B_{k\phi_j}^{v_j}=B_{k\phi_j}$ if $v_j=1$, and $B_{k\phi_j}^{v_j}=\lnot B_{k\phi_j}$ if $v_j=0$ and is analogous to the atomic predicate $\lnot \phi_j$. The cardinality of a bitmap index $|B_{k\phi_j}|$ is defined as the number of set bits in $B_{k\phi_j}$, for $j=1,\ldots,m$. An index is created for each pair of an atomic predicate and table column within $R$.

To compute the cardinality of any fragment $R_{(v_1,\ldots,v_m)}$ a "tuple-encoded" bitmap index is used: instead of encoding each atomic predicate by the domain of each attribute, the encoding is performed with respect to each tuple in the relation $R$. Only a single index is created for each atomic predicate indicating the table rows that satisfy the predicate. Estimating the cardinality of a fragment $R_{(v_1,\ldots,v_m)}$ breaks down to a calculation equivalent to Equation~\ref{eq:stadistinct_bitmap_index}. The index flavor will be further indicated by an upper-left superscript; $\mathcal{V}$ standing for "value-encoded" and $\mathcal{T}$ standing for a "tuple-encoded" bitmap index.

The number of disk pages for each table can be computed as easy as equating ratios of tuples to disk pages. This information, however, can be system specific and alternative DBMS-specific calculation approaches can be employed.

\section{Prediction of Horizontal Partitioning in PostgreSQL}
\label{sec:implementation_postgres}
The prediction of data partitioning is performed by simulation of horizontal partitioning. Here, we present an implementation in the PostgreSQL DBMS. PostgreSQL's query optimizer is used to estimate the total execution cost of a given workload across different partitioned schemas. The statistical data structures are completely compatible with PostgreSQL's catalogs.
Two PostgreSQL system catalogs are used to implement the simulation method: \textit{pg{\textunderscore}class} and \textit{pg{\textunderscore}statistic}. The \textit{pg{\textunderscore}class} catalogs tables, indexes, sequences, views and composite types~\cite{PGSQLCatalogs}, and \textit{pg{\textunderscore}statistic} stores statistical data about the contents of the database~\cite{PGSQLCatalogs}. The data loading stage merely consists of populating these two catalogs used by the query optimizer to construct a query execution plan. 
The fragment's data statistics are stored in \textit{pg\_statistic} and include one-dimensional histograms, most common attribute values, their frequencies and the width and domain cardinality of each attribute.

Most importantly, our approach is consistent with the PostgreSQL query optimizer because the catalog statistics that we leverage are exactly the ones used by the optimizer for query processing (more information regarding this can be found in~\cite{PGSQLDocumentationStatsEstimation}).

\subsection{Estimation of Statistical Data in pg\_statistic}
\label{sec:pg_statistic}
The histograms, most common values and frequencies arrays, the width and the fraction of distinct values for each fragment can be easily estimated from their counterparts for the parent relation, using existing statistics and partitioning predicate selectivities~\cite{PGSQLDocumentationStatsEstimation}. This estimation approach is very efficient (quick), but, on the other hand, it is very limited due to the assumptions of uniform data distribution and attribute independence (non-correlated attributes). These assumptions are rarely applicable in a real scenario, where highly correlated, non-uniform data occur very often. Thus, such derived statistics could be highly inaccurate and lead to wrong execution cost estimations. This approach is inextensible to other attributes that might or might not have predicates defined over them (the extracted set of predicates is defined over one or more attributes of each table in the non-partitioned schema).

Therefore, the histogram $H_{R_{(v_1,\ldots,v_m)}}$  and most common values $Val_{R_{(v_1,\ldots,v_m)}}$ for each fragment are derived exactly as described in Section~\ref{derived_stats}, while the multidimensional histogram of finest granularity $\mathit{MDH}_R$ is employed to efficiently compute the frequencies $Fre_{R_{(v_1,\ldots,v_m)}}$ of the most common values, as described in Section~\ref{sec:multidim_hist}. The width $w_k$ of each attribute $A_k$ of a given fragment $R_{(v_1,\ldots,v_m)}$ has exactly the same value as in $R$.  

PostgreSQL uses a signed $\sigma$ value to distinguish distinct value ratios against actual numbers of distinct values (domain cardinality). A positive value indicates the actual number of distinct values of the attribute (cardinality of its domain), and a negative value represents fraction that distinct values occupy in the in the relation~\cite{PGSQLCatalogs}, i.e $\sigma=-|Dom(A)|\,/\,|R|$ and has to be recalculated for the fragment $R_{(v_1, \ldots, v_m)}$.
Thus,
\begin{itemize}
	\item[1.] If $\sigma_R>0$, 
	\[
	\sigma_{R_{(v_1,\ldots,v_m)}} = |^{\mathcal{V}}B_{k\{\phi_1,\ldots,\phi_m\}}^{(v_1,\ldots,v_m)}|.
	\]
	\item[2.] If $\sigma_R<0$,
	\[
	\sigma_{R_{(v_1,\ldots,v_m)}} = -\frac{|^{\mathcal{V}}B_{k\{\phi_1,\ldots,\phi_m\}}^{(v_1,\ldots,v_m)}|}{|R_{(v_1,\ldots,v_m)}|},
	\]
	where $|R_{(v_1,\ldots,v_m)}|$ is the cardinality (number of tuples) of the fragment and can be efficiently computed by an equivalent bitmap operation $|R_{(v_1,\ldots,v_m)}|=|^{\mathcal{T}}B_{\{\phi_1,\ldots,\phi_m\}}^{(v_1,\ldots,v_m)}|$, where the bitmap index now encodes distinct tuples of $R$, rather than distinct values of $A_k$.
	\item[3.] If $\sigma_R=0$, then $\sigma_{R_{(v_1,\ldots,v_m)}}=0$, since the fraction of distinct values is not known~\cite{PGSQLCatalogs}.
\end{itemize}
In all cases, 1 through 3, it holds that
\[
|^{\mathcal{V}}B_{k\{\phi_1,\ldots,\phi_m\}}^{(v_1,\ldots,v_m)}| = |^{\mathcal{V}}B_{k\{\phi_1\}}^{(v_1)}\,\land\,^{\mathcal{V}}B_{k\{\phi_2\}}^{(v_2)}\,\ldots\,^{\mathcal{V}}B_{k\{\phi_m\}}^{(v_m)}|,
\]
\[
|^{\mathcal{T}}B_{\{\phi_1,\ldots,\phi_m\}}^{(v_1,\ldots,v_m)}| = |^{\mathcal{T}}B_{\{\phi_1\}}^{(v_1)}\,\land\,^{\mathcal{T}}B_{\{\phi_2\}}^{(v_2)}\,\ldots\,^{\mathcal{T}}B_{\{\phi_m\}}^{(v_m)}|.
\]

\subsection{Estimation of Statistical Data in pg\_class}

 The \textit{pg\_class} catalog stores physical storage information for each relation, such as the number of tuples and disk pages. The final step is to estimate the values of the fields \textit{reltuples} and \textit{relpages}. They have a unique value for each relation. The value of \textit{reltuples} for the fragment $R_{(v_1,\ldots,v_m)}$ is exactly the cardinality $|R_{(v_1,\ldots,v_m)}|$, i.e

\[
\mathit{reltuples}_R{(v_1,\ldots,v_m)} = |R{(v_1,\ldots,v_m)}|=|^{\mathcal{T}}B_{\{\phi_1,\ldots,\phi_m\}}^{(v_1,\ldots,v_m)}|.
\]

The value of \textit{relpages} represents the number of disk pages (physical blocks) that PostgreSQL uses to store the given relation. This value is sensitive to the total length (number of bytes) of each tuple, which depends on the data types used to represent the attributes of the relation.
If at least one attribute in \{$A_1,\ldots,A_n$\} is represented with a variable-length data type, then the most accurate estimate of \textit{replages} is
\[
\mathit{replages}_{R_{(v_1,\ldots,v_m)}} = \left \lceil \dfrac{|R_{(v_1, \ldots, v_m)}|}{\left \lfloor \dfrac{8168}{8+\sum_{i=1}^{n} w_i} \right \rfloor} \right \rceil.
\]
A disk page in PostgreSQL's  storage system is an abstraction layer over a physical block on disk. The default block size is 8KB. Each page contains a header of 24B, leaving 8168B free for tuples. Each tuple is associated with a 4B pointer to an array of offsets, 4B each, indicating the offset of each tuple stored on the page. Thus, a tuple requires a total of $8+\sum{i=1}^n w_i$ bytes of space, where $w_i$ is the calculated average width of $A_i$ in the fragment $R_{(v_1,\ldots,v_m)}$.

If all attributes of $R$ are represented with a fixed-length data type, then $\mathit{replages}_{R_{(v_1,\ldots,v_m)}}$ can be also computed as
\[
\mathit{replages}_{R_{(v_1,\ldots,v_m)}}=\left\lceil\frac{\mathit{relpages}_R \times |R_{(v_1,\ldots,v_m)}|}{|R\,|}\right\rceil.
\]
The existing field $\mathit{reltuples}_R$ is not used in these estimations since it could be outdated in the existing statistics because it is only updated by VACUUM, ANALYZE, and a few DDL commands~\cite{PGSQLCatalogs}.

The \textit{pg\_statistic} catalog is populated for each attribute \{$A_1,\ldots,A_n$\}, and the \textit{pg\_class} catalog is also populated for every generated partition of $R$ using the set of atomic predicates \{$\phi_1,\ldots,\phi_m$\}. No actual data is loaded into any of the partitions of $R$, and at this point, PostgreSQL's query optimizer can be used to estimate the total execution cost of set of queries $\mathcal{Q}$ from the given workload.

\subsection{The Simulation Process}
\label{simprocess}

The partitioning predicates for each relation $R$ are selected from $\mathcal{P}$ and non-overlapping check constraints are added for each partition of $R$ represent the range partition. 

The predicates for partitioning are extracted from the Star Schema Benchmark (SSB) workload (13 queries), and are subsequently split into atomic predicates. The supported predicate operators are: $>$, $\geq$, $<$, $\leq$, $=$, $\neq$, BETWEEN and IN. Predicates that contain the BETWEEN operator are split into two atomic predicates, while those that contain the IN operator are split into two or more atomic predicates. Each resulting atomic predicate represents a relation between an attribute and a constant. The number of resulting predicates depends on the number of elements in the IN clause.
For a set of atomic partitioning predicates \{$\phi_1,\ldots,\phi_m$\} over a relation $R$ in the non-partitioned schema and for each valid combination $(v_1,\ldots,v_m)$ , a partition is created. Its statistics are updated for each attribute in \textit{pg\_statistic}, and a single record is inserted into \textit{pg\_class}.

With the new partitioned schema in place, the process proceeds by estimation of the total execution cost of the selected queries from the workload. The total cost is the sum of the costs of all queries. If a given query does not involve a join operation, then its cost is estimated against all partitions of the relation on which the query is defined. On the other hand, if the query involves a join operation between $k$ relations, then its cost is estimated over every $k$-tuple of $k$ partitions that contains a nonempty intersection of the $k$ ranges by any attribute in the JOIN clause.

\section{Optimal Horizontal Schema Partitioning}
\label{optimal}

 The number of generated partitioned schemas grows exponentially as the number of predicates used for abstraction increases.
We want to compute an (near) optimal number of fragments such
that the performance of queries will be optimal. 
A GA is often used as an optimization approach in databases and data warehouses. An optimal design of a distributed database, in terms of query execution performance, can be subjected to a GA~\cite{distributeddbga}.
The GA approach allows to incorporate ad-hoc constraints to the optimization procedure, such as the maximal number of partitions that can be maintained by a data warehouse administrator, or a bounded data reallocation cost.
We now formally define the problem of finding an optimal partitioning implementation schema
of a data warehouse.

\subsection{The Optimization Problem}

 Let $(F,D_1, D_2, \ldots, D_k)$ be a star schema, 
$\mathcal{Q} = \{ Q_1, Q_2, \ldots, Q_l \}$ be a set of queries, and
$\mathsf{Cost}$ be a cost evaluation function.
The optimization problem of initial horizontal partitioning is defined as follows. Find a set of sub-star fragments $\mathcal{S}=\{ S_1, S_2, \ldots, S_N \}$ such that the cost
\[
\mathsf{MIN Cost}(\mathcal{S},\mathcal{Q}) \ 
\]
subject to the constraint $N \leq W$, where $W$ is a threshold representing
a maximal number of fragments that can be generated.

The optimization problem of horizontal re-partitioning is defined as follows. Find a set of $N$ sub-star fragments such that the cost
\[
\mathsf{MIN Cost}(\mathcal{S},\mathcal{Q}) \ 
\]
subject to the constraint $N \leq W$ and $L \leq WW$, where $WW$ is threshold representing
a maximal number of tuples (bytes) that can be relocated (read/written).

The cost evaluation function uses PostgreSQL's query optimizer to calculate the total execution cost of each particular solution (horizontally partitioned schema).
The cost of answering a query $Q_{i}$, denoted as $\mathsf{Cost}(\mathcal{S},Q_{i})$,
is equal to the value estimated by the optimizer.

\subsection{The Optimization Procedure}

 We now describe an optimization procedure for obtaining an optimal partitioning
implementation scheme given a workload:
\begin{description}
	\item[1] Extract all predicates $\mathcal{P}$ used by $\mathcal{Q}$.
	
	\item[2] Find a complete set of predicates $\mathcal{P}_i \subseteq \mathcal{P}$ ($1 \leq i \leq k$)
	corresponding to each dimension relation $D_i$.
	
	\item[3] Use $\textbf{ComputeMin}(\mathcal{P}_i,\mathcal{D}_i)$ procedure to find a minimal
	set of predicates for each $\mathcal{D}_i$. This procedure eliminates all redundant predicates
	in $\mathcal{P}_i$ which lead to no additional fragments.
	
	\item[4] Apply a genetic algorithm to find an optimal partitioning scheme.
\end{description}

The defined problem is an optimization problem and a GA is used to find an approximately optimal solution.
Candidate solutions to a given problem, also called \emph{chromosomes}, are
most commonly represented as bit strings, but other encodings are also possible.
The algorithm starts from a population of randomly generated solutions and proceeds in iterations (i.e.\ generations). At each generation, the cost of every solution
in the population is evaluated, multiple solutions are selected from the current
population based on their cost, and modified (recombined and possibly randomly mutated)
to form a new population. The new population is then used in the next iteration.
The algorithm terminates when either a maximum number of generations has been
produced, or a solution with satisfactory cost has been found. We now present
the design of our genetic algorithm. One of the choice factors for a GA is the flexibility due of the fitness function that is optimized (usually maximized). It allows one to add constraints of different nature and type that could fit and meet the needs of different environments and users, as well as constraints that come in the form od outside factors, or environmental factors. It allows a greater extent of adaptivity, widening the range of feasible adaptation requirements.

\subsection{Representation of the Solution}

 Let $\mathcal{P}_i = \{ \phi_{i,1}, \phi_{i,2}, \ldots \phi_{i,m_i} \}$ ($1 \leq i \leq k$)
be a complete and minimal set of predicates that needs to be applied to the dimension $D_i$
for horizontal partitioning. A possible solution of our problem
is a set of $N$ ($N \leq W$) different sub-star fragments. Each fragment $S_j$ ($1 \leq j \leq N$)
is represented by a bit array (or, bit-vector).
\[
(v_{1,1}, \ldots, v_{1,m_1}, \ldots, v_{k,1}, \ldots, v_{k,m_{k}})
\]
containing one bit for each predicate used in the partitioning.
Each bit in the solution is set to 1, if the respective predicate is satisfied
by all tuples in $S_j$; otherwise it is set to 0. So, we have that
\begin{flalign*}
S_j &= F_{(v_{1,1}, \ldots, v_{1,m_1}, \ldots, v_{k,m_{k}})} \\
&= \big( F \ltimes D_{1_{(v_{1,1}, \ldots, v_{1,m_1})}} \ltimes \ldots \ltimes D_{k_{(v_{k,1}, \ldots, v_{k,m_k})}} \big)
\end{flalign*}
The entry from the local index table pointing to $S_j$ will be its
bit array representation $(v_{1,1}, \ldots, v_{1,m_1}, \ldots, v_{k,1}, \ldots, v_{k,m_{k}})$.
In this way, we obtain that the search space of our optimization problem is
$2^{N \, \sum^{k}_{i=1}m_i}$, or in the worst case it is $2^{W \, \sum^{k}_{i=1}m_i}$.

A chromosome consists of $N$ composite genes, where each composite gene is a bit-vector representing one fragment $S_{j}$ as described above. One chromosome represents one possible solution to the problem.

\subsection{Genetic Algorithm Operators}

 A single point crossover operator is used, which chooses a random bit from two parent chromosomes, i.e.\ solutions, and then performs a swap of that bit and all subsequent
bits between the two parent chromosomes, in order to obtain two new offspring chromosomes.

The mutation operation is performed over each gene of a chromosome and mutates them with a given probability.
Because the genes are represented as bit arrays, a mutation of a gene means flipping the value of every bit with the given probability.

We use a natural selection operator where a chromosome is selected for survival in the next generation with a probability inversely proportional to the cost of the solution represented by the chromosome. A strategy of elitist selection is also used where the best chromosome of the population in the current generation is always carried unaltered to the population in the next generation.

The termination of the GA is established by restraining the number of generations evolved by the GA.

\section{Adaptivity}
\label{sec:adaptivity}

A key factor that influences our approach is adaptivity. The question of adaptivity is triggered when dealing with dynamically changing data or dynamically changing workloads over time. To ensure that the approach is adaptive to changes of this kind, it is crucial that the existing horizontally partitioned schema is able to re-adapt to the changes. A typical question induced from the presented statistics derivation strategies concerns their usability and efficiency when data changes over time (a dynamically-changing workload over time), or for short - the adaptivity of the approach. There are two key-points that constitute an answer. First and foremost, when data change, the statistical catalog data have to be re-created, followed by generation of partition-specific statistics for cost estimation. The latter requires re-creation of the statistical data structures (the multidimensional histogram and Roaring Bitmap indexes). This is a time-consuming operation in OLTP scenarios, but can be ameliorated by periodical updates of the data structures (updating an AVL tree or even lists is fast if implemented cleverly). Our approach is primarily targeted at OLAP systems in typical data warehousing scenarios, where data change occasionally, between longer periods of time - in contrast to the OLTP where changes occur on a regular basis. Re-creating the multidimensional histogram and bitmap indexes could be considered acceptable, if scheduled timely;  for example, in low-activity periods of reduced or negligible load (little or no update operations at all), such as weekends or holidays. Another viewpoint of adaptivity concerns maintenance, such as the maximal number of partitions that the data warehouse administrator (DWA) can maintain. This adaptivity to outside factors that are partially system-dependent can be achieved by extending the fitness function of the GA. Finally, the targeted scenario of our approach is offline re-partitioning.

\subsection{Adaptation to Dynamically Changing Workloads Over Time}
\label{adaptivity_workloads}
The first scenario that we consider in order to make our approach adaptive is characterized by changes in the workload . Specifically, a change of the workload is either adding a new query or removing an existing one. Either of these changes affects the fitness function of the GA (the total cost of the new workload), and more importantly, it introduces an additional \textit{cost of adaptation}. Removing a query form the workload is equivalent to removing all of its atomic predicates for partitioning. On the other hand, adding a query to the workload is equivalent to adding new atomic predicates for partitioning. The cost of adaptation is virtually the cost of re-partitioning the data regions affected by the changes. 

Assume that one round of partitioning has been already performed and there are change to the workload at that point. Let the existing (original) set of atomic predicates that have been previously used to approximately optimize the horizontally partitioned schema be denoted by $\mathcal{P}=\{\phi_q\,|\,\forall q\in\mathcal{Q}\}$, where $\mathcal{Q}$ is the original workload (set of queries). Let $\mathcal{P}_q\subset\mathcal{P}$ be the set of atomic predicates in query $q$. We distinguish two cases.

\begin{enumerate}
\item {\textbf{Query removal. } Suppose query $q$ is removed. This is the case when $\mathcal{Q}'=\mathcal{Q}\setminus \{q\}$, for some $q\in\mathcal{Q}$.  This yields the new set of atomic predicates $\mathcal{P}' = \mathcal{P}\setminus\mathcal{P}_q$. The adaptation task is to update the existing horizontally partitioned schema in order to reflect the changes in $\mathcal{Q}$ and $\mathcal{P}$, respectively. In other words, we adapt it to $\mathcal{Q}'$ and $\mathcal{P}'$. This is achieved by \textit{merging} all partitions that were generated by any $\phi_q$ or $\neg \phi_q$ alone, such that $\phi_q\in\mathcal{P}_q$. Then, for any relation $R$ over which $\phi_q$ is defined, for any fragment of $R$  generated by a set of predicates $\{\phi_1,\dots,\phi
_q,\dots,\phi_m\}$, we merge all fragments
\begin{equation}
R_{(v_1,\dots,v_{q-1},v_{q+1},\dots,v_{m})} = \bigcup_{v_q} R_{(v_1,\dots,v_{q-1},v_q,v_{q+1},\dots,v_m)},\quad R_{(.)}\subset R,
\label{eq:adaptivity_merge}
\end{equation}
i.e. all fragments generated by a set of predicates that contains $\phi_q$ are merged together.

This time, again, the PostgreSQL query optimizer is suitable to estimate the cost of merging since it involves the $\mathrm{SELECT/DELETE}$ and $\mathrm{UPDATE}$ SQL directives. Let $\mathsf{Cost_{mrg}}(R_i,R_j)$ denote the cost (in units interpreted as number of accesses to disk) to merge two tables (fragments/relations) $R_i$ and $R_j$, such that $i\neq j$. It is clear that the most efficient way to merge all fragments is to 'concatenate' them to the largest one (having the largest cardinality). Let
\[R_{(v_1,\dots,v_q,\dots,v_m)}^{max} = \argmax \limits_{R}\left[|R_{(v_1,\dots,v_q,\dots,v_q)}|\right],\quad v_q\in\{0,1\}.
\]
Then,
\begin{equation}
\mathsf{Cost_{mrg|\mathcal{P}',\mathcal{Q}'}} = \sum_{v_q}\mathsf{Cost_{mrg}}(R_{(v_1,\dots,v_q,\dots,v_m)}^{max},R_{(v_1,\dots,v_q,\dots,v_m)}),
\label{total_merge_cost}
\end{equation}
where $\mathsf{Cost_{mrg}}(R_{(v_1,\dots,v_q,\dots,v_m)}^{max},R_{(v_1,\dots,v_q,\dots,v_m)})=$ \textit{cost to read} $R_{(v_1,\dots,v_q,\dots,v_m)} +$ \textit{cost to write} $R_{(v_1,\dots,v_q,\dots,v_m)}$ \textit{to} $R_{(v_1,\dots,v_q,\dots,v_m)}^{max}$ .
}
\\

\item {\textbf{Adding a new query. } Similarly, we are now looking at splitting, rather than merging partitions. Assume that a new query $q_{new}$ is added to $\mathcal{Q}$, such that $\mathcal{Q}'=\mathcal{Q}\cup \{q_{new}\}$. Also, $\mathcal{P}'=\mathcal{P}\cup\mathcal{P}_{q_{new}}$. The strategy is strikingly similar to (1). In this case, for any $\phi_{q_{new}}$ defined over a relation $R$, we split any fragment of $R$ into two new complementary sub-fragments, such that
\begin{flalign*}
R_{(v_1,\dots,v_{q_{new}-1},v_{q_{new}+1},\dots,v_m)}&=R_{(v_1,\dots,v_{q_{new}-1},0,v_{q_{new}+1},\dots,v_m)} \\
&\cup R_{(v_1,\dots,v_{q_{new}-1},1,v_{q_{new}+1},\dots,v_m)},
\end{flalign*}
and
\[
R_{(v_1,\dots,v_{q_{new}-1},0,v_{q_{new}+1},\dots,v_m)} \cap R_{(v_1,\dots,v_{q_{new}-1},1,v_{q_{new}+1},\dots,v_m)}=\emptyset.
\]

The cost of splitting is equivalent to the cost of merging because it involves reading a fragment and then writing it to a new location on disk. Therefore, the most efficient solution again consists of comparing the cardinalities of $R_{(v_1,\dots,v_{q_{new}-1},0,v_{q_{new}+1},\dots,v_m)}$ and $R_{(v_1,\dots,v_{q_{new}-1},1,v_{q_{new}+1},\dots,v_m)}$, and then split by the one that has a lower cardinality, i.e. read and re-write the smaller sub-fragment. It is clear that the total cost is calculated in the same manner as in (1), i.e. it is the sum of the costs to split each existing fragment $R_{(v_1,\dots,v_{q_{new}-1},v_{q_{new}+1},\dots,v_m)}$.

}
\end{enumerate}

\subsection{Adaptation to Dynamically Changing Data Over Time}
\label{adaptivity_data}
The second scenario where adaptation to changes in necessary involves changes to the data, and not the workload. Usually, the worst case is when the size of a partition unexpectedly grows, resulting in unpredictable performance deterioration. If the partition is being accessed frequently by the given workload, it is clear that the total execution cost of the workload is neither small, nor optimal. In fact, the previously approximately optimal solution is now worse.

If this is the case, the statistical data for each partition have to be updated accordingly to reflect the most recent changes of data. Then, they are again eligible to be used by the query optimizer. This way, the consistency of the approach is preserved and the approach is adaptive to changes of the data. At this point, the GA is run again to search for a new approximately optimal horizontally partitioned schema. 

\section{Experimental Results}
\label{experiments}

 This section provides an experimental evaluation of the partitioning optimization approach. All experiments are conducted on a PostgreSQL 9.3 server, running under Ubuntu 15.04, that uses a single disk. All execution costs and times are measured using that disk as a reference point. The results presented in this section can be improved further by increasing the number of disks, which enables much of the queries to be distributed and parallelized across different partitions. The idea remains as an essential for our future work.

The SSB workload is used to find the optimal horizontally partitioned schema. The simulation method is implemented in Java, and the source code is available on-line at~\cite{gitsourcecode}. The JGAP genetic algorithm package is used to implement the GA.

The experiments are conducted on a generalized variation of the optimization problem, where classical, rather than derived horizontal partitioning is used, making the implementation suitable for data models other than a star schema, as well. The SSB workload is accompanied by 13 queries from which the predicates are extracted. They contain predicates for all relations in the schema. 

The experimental process consists of two stages:
\begin{itemize}
	\item[1.] Predictive optimization by the simulation method described in Section~\ref{prediction} using a GA with initial population size $K_{pop}$, evolved in $G$ generations. Additional GA parameters, such as the elitism $\epsilon$ and mutation probability $p_m$ are set at the initialization of the GA.
	\item[2.] Validation of the best solution at each generation. At each generation, the best solution found so far is re-evaluated on real data. The partitioned schema (solution) is re-created and actual data is inserted into all partitions. Then, VACUUM FULL ANALYZE is used to automatically generate the statistical data in catalogs for the partitioned schema. The cost total of all queries is estimated with the command EXPLAIN to validate the correctness of the our statistics estimation method, without executing any queries, but rather estimating their cost. Additionally, each query is then executed with EXPLAIN ANALYZE and the real cost and execution time are measured to validate the correctness and quality of the solution.
\end{itemize}

\subsection{Optimization of Horizontal Partitioning}

\begin{figure}[!h]
	\vspace{-0.2cm}
	\centering
	\includegraphics[width=0.6\textwidth]{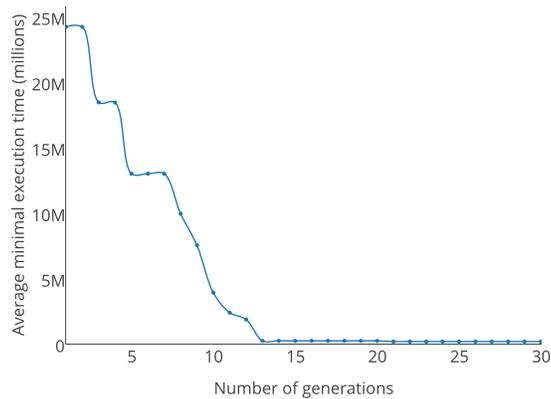}
	\caption{The average minimal estimated execution cost (in disk accesses) of SSB queries at each generation of the GA.}
	\label{fig:cost}
	\vspace{-0.1cm}
\end{figure}

 The first chart in Figure~\ref{fig:cost} shows the average minimal estimated execution cost (average best solution) by simulation of horizontal partitioning ($y$-axis) against the number of evolved generations ($x$-axis). The values are averaged over several runs of the GA. Each point on the line represents the total cost of the queries in the corresponding horizontally partitioned schema. The initial size of the population is $K_{pop}=20$, and $G=30$. The elitism is set to $\epsilon=2$, while the mutation probability is set to $p=0.1$. The execution cost in PostgreSQL is measured by five specific units of relative cost of different operations, and the actual value of any cost closely resembles the number of accesses to disk of the operation (query). 
 For brevity, let the cost represent the number of disk accesses required.

The results in Figure~\ref{fig:cost} show that the horizontal partitioning can be optimized significantly. The total execution cost across different partitioned schemas is reduced from 25 millions to less than 200,000 disk accesses, on average. The total execution cost in the non-partitioned schema is approximately 2 millions, so its is reduced more than 10 times in terms of disk accesses, on average.

\begin{figure}[!h]
	\vspace{-0.2cm}
	\centering
	\includegraphics[width=0.6\textwidth]{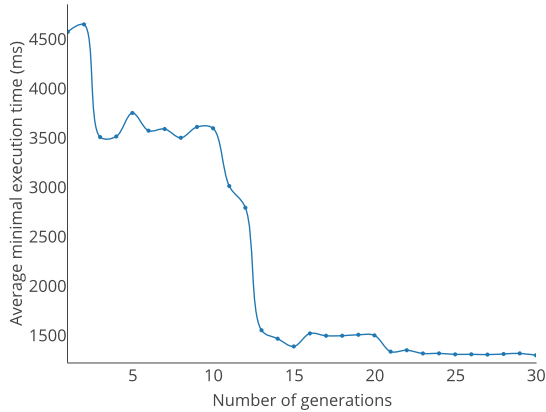}
	\caption{The average optimal total execution time of the SSB queries at each generation of the GA.}
	\label{fig:time}
	\vspace{-0.1cm}
\end{figure}

Figure~\ref{fig:time} shows the quality of the best solution at each generation, in terms of the average of the real total execution time of the queries.  This experiment is part of the validation stage and the total execution time of the workload is measured in milliseconds (ms), on the $y$-axis, at the current generation, shown on the $x$-axis. At each generation, the actual data are loaded into the partitions and queries are executed against the data.

The reduction of the total execution time follows the trend of reduction of the estimated workload cost in Figure~\ref{fig:cost}. The results indicate that our proposed approach is, in fact, suitable for real applications in time-critical scenarios. The total execution time of the queries across different partitioned schemas is reduced to approximately three times by the GA, on average. The total execution time of the queries in the non-partitioned schema is 11168 ms, while the average best solution reduces this time to approximately 1270 ms, or approximately 8.8 times, on average. Most importantly, these reduction are achieved on the same disk, solely by partitioning.

\subsection{Validation and Estimation Error Rate}
Finally, we validate and confirm the correctness of the approach by measuring the error rate of the estimated total execution cost of the solution at each generation. The results show that the approach is highly accurate, hence a multi-line plot of the execution cost is not suitable for visualization. The mean error rate in 30 generations is only 1.16\%$\pm$0.45\%, and the smallest and largest error rates are 0\% and 3.10\%, respectively. The low error rate confirms correctness of the statistics estimation method in our approach.

\section{Related Work}
\label{related}
Automatic partitioning in commercial systems has already been a practice. Automatic skew-aware partitioning has been targeted at enterprise On-Line Transaction Processing (OLTP) systems (a feature of the shared-nothing H-Store)~\cite{pavlo2012}. However, in terms of consistency and efficient query distribution, its challenges are different than partitioning in On-Line Analytical Processing (OLAP) systems, at which this paper is targeted. Thus, the relationship between the two is obscure, because the challenge here is optimality rather than consistency.

\section{Conclusion and Future Work}
\label{conclusion}

 In this paper we describe a novel approach for predictive horizontal partitioning, based on a formal model for partitioning by predicate abstraction. We demonstrate how this approach can be used to find an optimal data warehouse design that improves the performance of the system for a given workload of data and queries. The advantage of our approach is the simulation method based on estimation of the database statistics, for which loading of any real data or query execution is unnecessary. The latter is attained by using a real query optimizer. The experimental evaluation in the last section confirms that the approach is applicable to real systems and provides a clear prospect of its contribution to system performance improvement.  

The next possible direction of extension of our approach is to evaluate its performance on distributed systems with multiple nodes, where distributed queries provide the opportunity for parallel execution, which will emphasize the minimization of the workload execution time. The approach will be tested on different number of nodes, several scaling factors of the SSB dataset, and different constraints of the optimization problem, such as the maximal number of partitions, partition maintenance cost and other environmental factors.

\bibliographystyle{splncs03}
\bibliography{refs.bib}

\begin{thebibliography}{10}
\providecommand{\url}[1]{\texttt{#1}}
\providecommand{\urlprefix}{URL }

\bibitem{gitsourcecode}
{Predicate Partitioning Repository}.
  \url{https://bitbucket.org/predpart/predicate\_partitioning} (2015),
  accessed: 2015-11-12

\bibitem{Akdere:2012:LQP:2310257.2310339}
Akdere, M., Çetintemel, U., Riondato, M., Upfal, E., Zdonik, S.B.:
  Learning-based query performance modeling and prediction. In: Proceedings of
  the 2012 IEEE 28th International Conference on Data Engineering. pp.
  390--401. ICDE '12, IEEE Computer Society, Washington, DC, USA (2012),
  \url{http://dx.doi.org/10.1109/ICDE.2012.64}

\bibitem{bellatreche2013horizontal}
Bellatreche, L., Bouchakri, R., Cuzzocrea, A., Maabout, S.: Horizontal
  partitioning of very-large data warehouses under dynamically-changing query
  workloads via incremental algorithms. In: Proceedings of the 28th Annual ACM
  Symposium on Applied Computing. pp. 208--210. ACM (2013)

\bibitem{bellatrechereferential}
Bellatreche, L., Boukhalfa, K., Richard, P., Woameno, K.Y.: Referential
  horizontal partitioning selection problem in data warehouses: Hardness study
  and selection algorithms. International Journal of Data Warehousing and
  Mining (IJDWM)  5(4),  1--23 (2009)

\bibitem{LemireRoaringBitmaps}
Chambi, S., Lemire, D., Kaser, O., Godin, R.: Better bitmap performance with
  roaring bitmaps. Software: Practice and Experience  (2015)

\bibitem{dimovski2010horizontal}
Dimovski, A., Velinov, G., Sahpaski, D.: Horizontal partitioning by predicate
  abstraction and its application to data warehouse design. In: Advances in
  Databases and Information Systems, Lecture Notes in Computer Science, vol.
  6295, pp. 164--175. Springer Berlin Heidelberg (2010),
  \url{http://dx.doi.org/10.1007/978-3-642-15576-5\_14}

\bibitem{Ganesan:2004:OBR:1316689.1316729}
Ganesan, P., Bawa, M., Garcia-Molina, H.: Online balancing of range-partitioned
  data with applications to peer-to-peer systems. In: Proceedings of the
  Thirtieth international conference on Very large data bases - Volume 30. pp.
  444--455. VLDB '04, VLDB Endowment (2004),
  \url{http://dl.acm.org/citation.cfm?id=1316689.1316729}

\bibitem{Herodotou:2010:XSQ:1920841.1920984}
Herodotou, H., Babu, S.: Xplus: a sql-tuning-aware query optimizer. Proc. VLDB
  Endow.  3(1-2),  1149--1160 (Sep 2010),
  \url{http://dl.acm.org/citation.cfm?id=1920841.1920984}

\bibitem{DBLP:conf/birte/JindalD11}
Jindal, A., Dittrich, J.: Relax and let the database do the partitioning
  online. In: BIRTE. pp. 65--80 (2011)

\bibitem{61401488}
Ke, Q., Prabhakaran, V., Xie, Y., Yu, Y., Wu, J., Yang, J.: {Optimizing Data
  Partitioning for Data-Parallel Computing}  (2011)

\bibitem{DBLP:dblpconf/dexa/LirozGistauAPPV12}
Liroz-Gistau, M., Akbarinia, R., Pacitti, E., Porto, F., Valduriez, P.,
  Valduriez, P.: Dynamic workload-based partitioning for large-scale databases.
  In: DEXA (2). pp. 183--190 (2012)

\bibitem{JGAP}
Meffert, K.: {Java Genetic Algorithms and Genetic Programming Package.}
  \url{http://jgap.sf.net} (2015), accessed 2015-25-11

\bibitem{Nehme:2011:APD:1989323.1989444}
Nehme, R., Bruno, N.: Automated partitioning design in parallel database
  systems. In: Proceedings of the 2011 ACM SIGMOD International Conference on
  Management of data. pp. 1137--1148. SIGMOD '11, ACM, New York, NY, USA
  (2011), \url{http://doi.acm.org/10.1145/1989323.1989444}

\bibitem{OracleDB}
{Oracle Database}: {Database SQL Tuning Guide: Histograms}.
  \url{https://docs.oracle.com/database/121/TGSQL/tgsql\_histo.htm\#TGSQL366}
  (2015), accessed: 2015-11-02

\bibitem{Ozsu2011principles}
{\"O}zsu, M.T., Valduriez, P.: Principles of distributed database systems.
  Springer Science \& Business Media (2011)

\bibitem{o2007star}
O’Neil, P.E., O’Neil, E.J., Chen, X.: The star schema benchmark (ssb). Pat
  (2007)

\bibitem{Pavlo:2012:SAD:2213836.2213844}
Pavlo, A., Curino, C., Zdonik, S.: Skew-aware automatic database partitioning
  in shared-nothing, parallel oltp systems. In: Proceedings of the 2012 ACM
  SIGMOD International Conference on Management of Data. pp. 61--72. SIGMOD
  '12, ACM, New York, NY, USA (2012),
  \url{http://doi.acm.org/10.1145/2213836.2213844}

\bibitem{pavlo2012}
Pavlo, A., Curino, C., Zdonik, S.: Skew-aware automatic database partitioning
  in shared-nothing, parallel {OLTP} systems. In: SIGMOD '12: Proceedings of
  the 2012 international conference on Management of Data. pp. 61--72 (2012),
  \url{http://hstore.cs.brown.edu/papers/hstore-partitioning.pdf}

\bibitem{PGSQLCatalogs}
PostgreSQL: {Documentation Chapter 47 - System Catalogs}.
  \url{http://www.postgresql.org/docs/9.3/static/catalogs.html} (2015),
  accessed: 2015-08-27

\bibitem{PGSQLDocumentationStatsEstimation}
PostgreSQL: {Documentation Chapter 60 - How the Planner Uses Statistics}.
  \url{http://www.postgresql.org/docs/9.3/static/planner-stats-details.html}
  (2015), accessed: 2015-10-25

\bibitem{sedgewick2002algorithms}
Sedgewick, R.: Algorithms in Java, Parts 1-4. Addison-Wesley Professional
  (2002)

\bibitem{distributeddbga}
Sevinç, E., Coşar, A.: Distributed database design with genetic algorithm and
  relation clustering heuristic. In: Gelenbe, E., Lent, R., Sakellari, G.,
  Sacan, A., Toroslu, H., Yazici, A. (eds.) Computer and Information Sciences,
  Lecture Notes in Electrical Engineering, vol.~62, pp. 133--136. Springer
  Netherlands (2010), \url{http://dx.doi.org/10.1007/978-90-481-9794-1\_27}

\bibitem{Stonebraker:2010:MPD:1629175.1629197}
Stonebraker, M., Abadi, D., DeWitt, D.J., Madden, S., Paulson, E., Pavlo, A.,
  Rasin, A.: Mapreduce and parallel dbmss: friends or foes? Commun. ACM  53(1),
   64--71 (Jan 2010), \url{http://doi.acm.org/10.1145/1629175.1629197}

\bibitem{wu13prediction}
Wu, W., Chi, Y., Zhu, S., Tatemura, J., Hac{\i}g\"{u}m\"{u}\c{s}, H., Naughton,
  J.F.: Predicting query execution time: are optimizer cost models really
  unusable? In: Proceedings of the 29th International Conference on Data
  Engineering. IEEE Computer Society (2013)

\end{thebibliography}
\end{document}